\documentclass[10pt,twocolumn,letterpaper]{article}

\usepackage{cvpr}              

\definecolor{cvprblue}{rgb}{0.21,0.49,0.74}
\usepackage[pagebackref,breaklinks,colorlinks,allcolors=cvprblue]{hyperref}

\def\paperID{*****} 
\def\confName{CVPR}
\def\confYear{2025}

\title{Lightweight Models for Emotional Analysis in Video}

\author{Quoc-Tien Nguyen, Hong-Hai Nguyen, Van-Thong Huynh\thanks{Corresponding author}\\
Dept. of ITS, FPT University\\
Ho Chi Minh City, 71216, Vietnam\\
{\tt\small \{tiennq27,hainh51,thonghv4\}@fe.edu.vn}
}

\begin{document}
\maketitle
\begin{abstract}


In this study, we present an approach for efficient spatiotemporal feature extraction using MobileNetV4 and a multi-scale 3D MLP-Mixer-based temporal aggregation module. MobileNetV4, with its Universal Inverted Bottleneck (UIB) blocks, serves as the backbone for extracting hierarchical feature representations from input image sequences, ensuring both computational efficiency and rich semantic encoding. To capture temporal dependencies, we introduce a three-level MLP-Mixer module, which processes spatial features at multiple resolutions while maintaining structural integrity. Experimental results on the ABAW 8th competition demonstrate the effectiveness of our approach, showing promising performance in affective behavior analysis. By integrating an efficient vision backbone with a structured temporal modeling mechanism, the proposed framework achieves a balance between computational efficiency and predictive accuracy, making it well-suited for real-time applications in mobile and embedded computing environments.
\end{abstract}    
\section{Introduction}
\label{sec:intro}

Human emotion recognition is a subfield of human behavior analysis that has got significant interest from researchers in artificial intelligence, psychology, and human-computer interaction. By utilizing various types of data, such as audio, images, and text, researchers can analyze and predict human emotions as well as continuous actions. Advancements in deep learning has significantly improved the accuracy and efficiency of emotion recognition systems. These technologies can detect emotions in real time, allowing applications across a wide range of fields. Understanding human emotions and behavior can create various applications in areas such as healthcare, autonomous vehicles, robotics, and more \cite{kollias2022abaw, haque2018measuring, sikander2018driver, kolhatkar2020sfu}. For example, emotion recognition is increasingly being used in mental health monitoring, where it helps identify signs of depression or stress based on behavioral patterns. In marketing, consumer emotion recognition enables brands to tailor their products and advertising strategies to evoke desired emotional responses. Moreover, in human-computer interaction, emotion-aware systems are enhancing user experiences by adapting to users’ emotional state.

The Emotion recognition has many benefits in life, but emotion recognition still faces various challenges in real-world applications. Human emotions are complex and influenced by various factors such as gender, age, and context, etc. Therefore, the 8th Affective \& Behavior Analysis in-the-Wild (ABAW8) workshop \cite{kollias2025advancements} specifically addresses the complex challenges inherent in analyzing human affective states and behavioral patterns in unconstrained, real-world scenarios. Unlike controlled laboratory environments, in-the-wild settings present numerous variables including diverse lighting conditions, varying head poses, occlusions, and spontaneous expressions that significantly complicate accurate analysis. 

The challenge includes six tasks: Valence-Arousal (VA) Estimation, Expression (EXPR) Classification, Action Unit(AU) Detection, Compound Expression (CE) Recognition, Emotional Mimicry Intensity (EMI) Estimation, Ambivalence/Hesitancy (AH) Recognition. The challenges utilize the Aff-Wild2 \cite{zafeiriou2017aff,kollias20247th,kollias20246th,kollias2024distribution,kollias2023abaw2,kollias2023multi,kollias2023abaw,kollias2022abaw,kollias2021analysing,kollias2021affect,kollias2021distribution,kollias2020analysing,kollias2019expression,kollias2019deep,kollias2019face}, C-EXPR-DB \cite{kollias2023multi}, HUME-Vidmimic2, and BAH datasets, providing a comprehensive evaluation framework for affective behavior analysis methodologies. In this paper, we solve the tasks such as Valence-Arousal (VA) Estimation, Action Unit(AU) Detection, Emotional Mimicry Intensity (EMI) Estimation, Ambivalence/Hesitancy (AH) Recognition. The VA task predicts valence and arousal in video sequences, while the AU task detects the presence of 12 action units (AUs) in each video frame. The EMI task estimates the intensity of six emotional dimensions (Admiration, Amusement, Determination, Empathic Pain, Excitement, and Joy). Lastly, the AH task identifies the presence or absence of ambivalence or hesitancy in each frame.




\section{Method}\label{sec:method}

\subsection{Visual feature extraction}

The visual feature extraction process is facilitated by a small variant of MobileNetV4~\cite{qin2024mobilenetv4}, pretrained on AffectNet dataset~\cite{mollahosseini2017affectnet}. MobileNetV4~\cite{qin2024mobilenetv4} introduces a universally efficient architecture optimized for mobile devices, integrating the Universal Inverted Bottleneck (UIB) and Mobile Multi-Query Attention (Mobile MQA) blocks to enhance feature extraction efficiency. The UIB block unifies key architectural components, including Inverted Bottleneck (IB), ConvNext, Feed Forward Network (FFN), and an Extra Depthwise (ExtraDW) variant, providing flexibility in spatial and channel mixing while improving computational efficiency. Mobile MQA accelerates attention mechanisms by over 39\% on mobile accelerators, further optimizing feature representation. Additionally, an advanced neural architecture search (NAS) strategy refines MobileNetV4’s design, ensuring mostly Pareto-optimal performance across CPUs, DSPs, GPUs, and dedicated accelerators like Google's EdgeTPU. 
In this study, we fed a sequence of $224\times224\times3$ images into the MobileNetV4 to extract multi-scale feature maps from input frames. The backbone is configured to output hierarchical feature representations at different spatial resolutions, allowing for rich semantic extraction. To maintain efficiency while preserving important spatial details, all but the final layers of the backbone are frozen during training. The extracted feature maps serve as input to the subsequent temporal modeling module.

\subsection{Temporal aggregation module}
To model temporal dependencies in the extracted features, we incorporate a multiscale 3D MLP-Mixer-based~\cite{tolstikhin2021mlp} to build temporal aggregation module (TAM). This module processes the sequential feature maps using multiple levels of spatial granularity. TAM consist of three mixer layers operate on different feature resolutions: (1) a high-resolution mixer for $28\times28$ feature maps, (2) a mid-resolution mixer with input size of $14\times14$, and (3) a low-resolution mixer for highly detail feature maps of size $7\times7$. Each mixer employs 3D MLP-based transformations to capture temporal relationships while preserving spatial structure. Finally, a fully connected layer maps the aggregated feature representations to the target output space, ensuring effective sequence-level prediction.

\section{Experiments and Results}\label{sec:exp_result}
\subsection{Dataset}
This subsection provides a brief summary of the datasets used in each challenge, such as Action Unit (AU) Detection, Valence-Arousal (VA) Estimation, Emotional Mimicry Intensity (EMI) Estimation.
\paragraph{Action Unit (AU) Detection}
This challenge uses a data set that includes 542 videos with annotations for 12 Action Units (AU), which represent facial muscle movements, including brow raisers, cheek raisers, lip tighteners, and jaw drops. The data set consists of 2,627,632 frames captured from 438 unique subjects. Annotations were developed through a semiautomatic methodology that integrates both manual and computational techniques. The data set has been divided into three subsets: a training set (295 videos), a validation set (105 videos), and a testing set (142 videos). \autoref{tab:au_dis} provides the distribution of the AU annotations of the dataset.


\paragraph{Valence-Arousal (VA) Estimation}
This challenge utilizes an expanded version of the Aff-Wild2 database, including 594 videos annotated for valence and arousal. The data set consists of 2,993,081 frames captured from 584 subjects. Significantly, sixteen videos contain dual subjects, both of whom received independent annotations. Four expert annotators evaluated the data set following the methodology detailed in \cite{cowie2000feeltrace}, continuous valence, and arousal values within the range of [-1, 1].

To maintain subject independence across experimental protocols, the data set has been partitioned into three discrete subsets: a training set (356 videos), a validation set (76 videos), and a testing set (162 videos). This division ensures that individual subjects appear exclusively in one subset.
\paragraph{Emotional Mimicry Intensity (EMI) Estimation.}

 The EMI Challenge considers emotional mimicry through the HUME-Vidmimic2 dataset, which includes more than 30 hours of audiovisual recordings from 557 participants. This data set was collected in naturalistic environments where participants used their webcams to mimic facial and vocal expressions presented in seed videos, subsequently self-evaluating their mimicry performance on a scale of 0-100.
The data set has been systematically partitioned according to the distribution detailed in \autoref{tab:HUME_stat}, which presents comprehensive statistics for each subset. To facilitate analysis, participants are provided with facial detections extracted from videos using MTCNN \cite{zhang2016joint}, processed at a sampling rate of 6 frames per second.

\begin{table}
    \centering
    \caption{HUME-Vidmimic2 partition statistics.
    \label{tab:HUME_stat}}
    \begin{tabular}{c c c} \toprule
        Partition &  Duration & \#Samples \\ 
        & (HH:MM:SS) & \\ \midrule
        Train & 15:07:03  & 8072 \\
        Validation & 9:12:02 & 4588  \\
        Test & 9:04:05  & 4586 \\
        \bottomrule
    \end{tabular}
\end{table}

Furthermore, to enable the development of end-to-end methodological approaches \cite{tzirakis2017end,tzirakis2018end2you,tzirakis2018end,tzirakis2021end,tzirakis2021speech}, participants receive pre-extracted features derived from the raw audiovisual data, specifically: facial features processed using Vision Transformer (ViT) \cite{caron2021emerging}, audio signals processed using Wav2Vec 2.0 \cite{baevski2020wav2vec}.

\paragraph{ Ambivalence/Hesitancy Recognition}


\paragraph{Action Unit (AU) Detection}
The $F1$ score uses precision and recall to calculate which ensure a robust assessment of classification performance. The equation of the $F_1$ score is as follows:

 \begin{equation}
    \begin{split}
        \label{eqn:F1_score}
        {F_1} &= \frac{2*precision*recall}{precision+recall}
    \end{split}
\end{equation}

The average F1 score is used to evaluate 12 AUs. The F1 score ranges from 0 to 1, where 1 represents perfect and 0 represents the worst performance. The formula is expressed as follows:
 \begin{equation}
    \begin{split}
        F_{\mathrm{AU}} &= \frac{\sum_{au}F_{1}^{au}}{12}
    \end{split}
\end{equation}

\paragraph{Valence-Arousal (VA) Estimation} In this task, the Concordance Correlation Coefficient (CCC) \cite{lawrence1989concordance}, $\mathcal{P}$ is used to evaluate the performance of the model. The CCC is a measure that is used to evaluate the agreement between two continuous variables. The CCC range is from [-1,1] where -1 is a negative correlation, 0 is no correlation, and 1 is a high correlation. The formula is expressed as follows:

 \begin{equation}
    \begin{split}
        \label{eqn:va_combined}
        \mathcal{P}_{\mathrm{VA}} &= \frac{\mathcal{P}_\mathrm{V} + \mathcal{P}_{\mathrm{A}}}{2}
    \end{split}
\end{equation}
where $\mathcal{P}_\mathrm{V}$ and $\mathcal{P}_\mathrm{A}$ are the CCC of valence and arousal, respectively, which is defined as
\begin{equation}
    \begin{split}
        \label{eqn:ccc}
        \mathcal{P} &= \frac{2 \rho \sigma_{\hat Y} \sigma_{Y} }{\sigma_{\hat Y}^2 + \sigma_{Y}^2 + (\mu_{\hat Y} - \mu_{Y})^2}
    \end{split}
\end{equation}

where $\mu_{Y}$ was the mean of the label $Y$, $\mu_{\hat Y}$ was the mean of prediction $\hat Y$, $\sigma_{\hat Y}$ and $\sigma_{Y}$ were the corresponding standard deviations,  $\rho$ was the Pearson correlation coefficient between $\hat Y$and $Y$. 

\paragraph{Emotional Mimicry Intensity (EMI) Estimation.}
The average Pearson’s Correlation Coefficient ($\rho$)is used to measure the six emotion dimensions:
 \begin{equation}
    \begin{split}
        \mathcal{P}_{\mathrm{EMI}} &= \frac{\sum_{i=1}^{6}{\rho}^{i}}{6}
    \end{split}
\end{equation}



\autoref{tab:au_task} shown results of our approach compared to previous studies on Affwild2 dataset~\cite{kollias2021affect}.
\begin{table}[htbp]
    \centering
    \caption{Evaluation on Action Unit detection task with F1 score.}
    \label{tab:au_task}
    \begin{tabular}{l c c}\toprule
       Methods  & Validation & Test \\ \midrule
       Regnet-ViT~\cite{vu2023ensemble} & 0.5280 & - \\ 
       EmotiEffNet~\cite{savchenko2024hsemotion,savchenko2025hsemotion} & 0.537 & 0.488 \\ 
       AIWELL-UOC~\cite{cabacasmaso2025} & 0.451 & 0.417 \\ \midrule
         Ours with 3 level mixer & 0.5369 & - \\
         Ours with 2 level mixer & 0.5338 & - \\
         Ours - single mixer & 0.5441 & 0.484\\
         \bottomrule
    \end{tabular}
\end{table}

{
    \small
    \bibliographystyle{ieeenat_fullname}
    \bibliography{main}
}


\end{document}